# Modelling daily mobility using mobile data traffic at fine spatiotemporal scale

Panayotis Christidis[1], María Vega- Gonzalo, Miklos Radics

*European Commission, Joint Research Centre, Seville Spain*

**Abstract**

We applied a data-driven approach that explores the usability of the NetMob 2023 dataset in modelling mobility patterns within an urban context. We combined the data with a highly suitable external source, the ENACT dataset, which provides a 1 km x 1km grid with estimates of the day and night population across Europe. We developed three sets of XGBoost models that predict the population in each 100m x 100m grid cell used in NetMob2023 based on the mobile data traffic of the 68 online services covered in the dataset, using the ENACT values as ground truth. The results suggest that the NetMob 2023 data can be useful for the estimation of the day and night population and grid cell level and can explain part of the dynamics of urban mobility.

## 1. Introduction

The use of mobile phone activity data for the analysis of human behaviour is a growing research field with numerous applications in mobility, health, socio-economic and demographic analyses (e.g. (Blondel et al., 2015; Deville et al., 2014; Gao et al., 2020; Li et al., 2019; Pappalardo & Simini, 2018; Stachl et al., 2020; Widhalm et al., 2015)). The majority of applications so far were based on Call Detail Records (CDR) data, which correspond to voice call and sms events. Such data can be sparse and irregular in time, leading to limitations in terms of the scope and scalability of applications.

The dramatic increase in the use of modern 4G cellular networks has changed the use of mobile phones, shifting towards a highly intense use of wireless network services. Such services generate significant volumes of data traffic and- in principle- could provide a larger spatio-temporal coverage than simple voice or sms traffic. The NetMob 2023 dataset (Martínez-Durive et al., 2023) is the first open dataset that provides detailed records on mobile data traffic. Through an innovative approach, the data provider (Orange) allocated the traffic through its base stations to a highly granular map of mobile traffic activity in 20 major cities in France. In addition, the activity is broken down into 68 main mobile applications, for both uploads and downloads.

As part of the NetMob 2023 Conference Data Challenge, we explore how this dataset can be used for the analysis of urban population dynamics and human mobility. We extended the methods used in the past on similar datasets provided- among others- by the same mobile phone operator (Christidis et al., 2022) in order to test how this new dataset can be combined with demographic data. The goal was to explore how the changes in mobile phone data activity can be used as a proxy for the dynamics of human activity in an urban context, in the line of what has been proposed by (Bwambale et al., 2019). We identified a highly suitable external dataset that can be used as the basis for the analysis, namely the ENACT dataset (Batista e Silva et al., 2020), a 1 km x 1km grid that provides estimates of the day and night population across Europe. We applied an XGBoost model that predicts the night and day population in ENACT using the data in the NetMob 2023 dataset, in order to evaluate the degree at which mobile phone data traffic can be used to estimate population in each cell.

---

[1] Corresponding author, e-mail: Panayotis.Christidis@ec.europa.eu

## 2. Data sources

### 2.1 The NetMob 2023 dataset

A full description of the NetMob 2023 dataset is available in (Martínez-Durive et al., 2023). The data consist of mobile data traffic volumes for 20 major cities in France, split into 100 x 100 m cells at 15 minute intervals, for the period 16 March 2019 to 31 May 2019 (77 days in total). Data traffic volumes are available for 68 different mobile services, for uploads and downloads, which represent more than 70% of total data traffic volume in France. The dataset is divided in individual files for each combination of city- mobile service- calendar day –traffic direction, with a total of 20x68x77x2= 209440 files. Each file consists of i rows (where i is the number of 100 x 100m cells in the corresponding city) and 96 columns (each representing a 15 minute interval within the corresponding calendar day) with the volume of data traffic $M_a^i(t)$, where $a$ denotes the mobile service, i the grid cell and t the time interval. Separate files are provided for uploaded and downloaded traffic volumes.

The total size of the dataset exceeds 4 Tb. The dataset requires a few data cleaning operations, mainly to account for the daylight saving time change on 31 March 2019, when clocks were shifted from 02:00 to 03:00, as well as for the treatment of outliers and anomalies (e.g. outages across the network or for specific internet services). In addition, NetMob 2023 provides the necessary data for the geographic representation of each city's dataset, through a GeoJson file for with the WGS84 coordinate system. Each GeoJson file contains the tile identifier of each regular grid tile in the target urban area and the corresponding polygon that bounds the 100×100 m tile.

The data for each mobile service at daily level show high variability in both spatial and temporal terms. Activity levels tend to vary significantly through the day and among tiles. The values at 15' intervals show abrupt oscillations, probably due to noise in the measurements, but follow more stable patterns when aggregated along larger periods (e.g. at 1 or 2 hour intervals). Most importantly, even though noise levels appear to be high for individual tiles in specific time slots/ days, the mean values across all 77 days of observations become more stable and indicate a high level of predictability.

Calculating the mean 1-hour time slot total for each weekday along the 11 weeks of data available already allows several observations concerning the spatiotemporal patterns of each application use. Figures 1a to 1f show –in logarithmic scale- the grid cells with the highest concentration of download traffic for Uber, Waze and Netflix for the 9:00-10:00 and 19:00-20:00 periods on Mondays (i.e. the mean of the four 15' periods across all seven Mondays in the dataset). The three example already demonstrate how the spatial distribution of the activity varies significantly among services and along the day. As in all 68 services analysed, the intensity of use of each service depends on the day of the week and the time of the day, as well as on the profile of each user. For example, an office employee, a student or a pensioner would all have a different mobile data activity profile at different moments during the day, that would be further modified if they were at or outside home, or in transit. In addition, the distribution at grid level of the number of users –in total and in terms of user profile- is highly variable. As a result, there is a high variance in the data traffic volume measured in individual grid cells in specific time periods. There is, nevertheless, a clear formation of clusters when measuring activity along the seven weeks covered in the data. The fact that those clusters differ by mobile service, day of the week or time suggests that these patterns are not directly proportional to the population within each grid cell, an observation that can be used as the basis for the analysis of the population dynamics that these patterns may reveal.

Fig.1a: Concentration of Uber use, Mondays 9:00-10:00

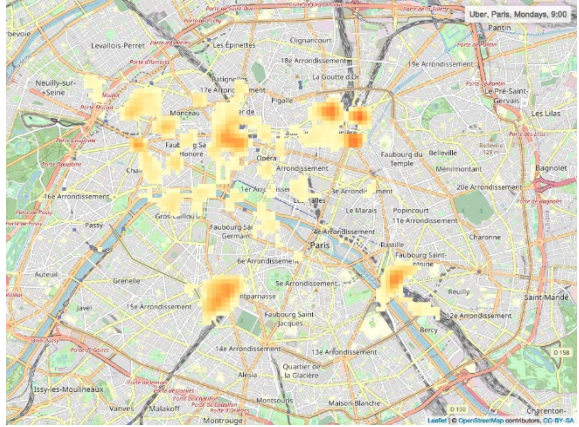

Fig.1b: Concentration of Uber use, Mondays 19:00-20:00

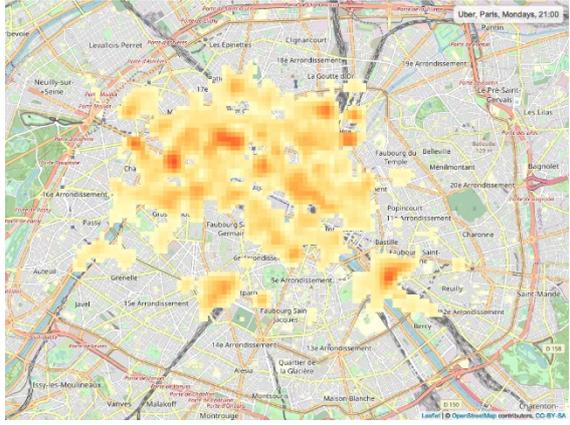

Fig.1c: Concentration of Waze use, Mondays 9:00-10:00

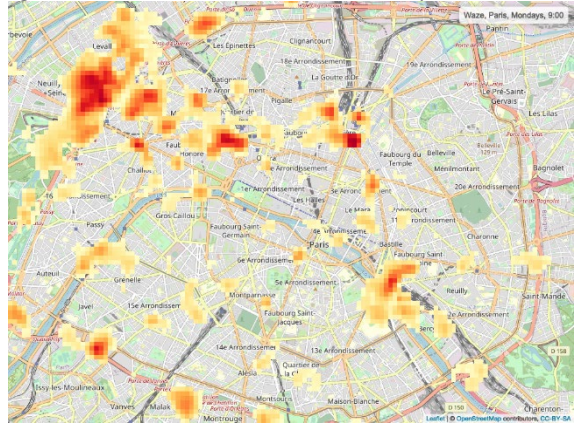

Fig.1d: Concentration of Waze use, Mondays 19:00-20:00

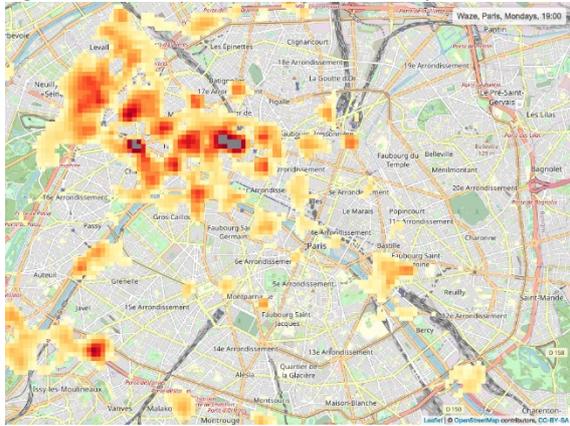

Fig.1e: Concentration of Netflix use, Mondays 9:00-10:00

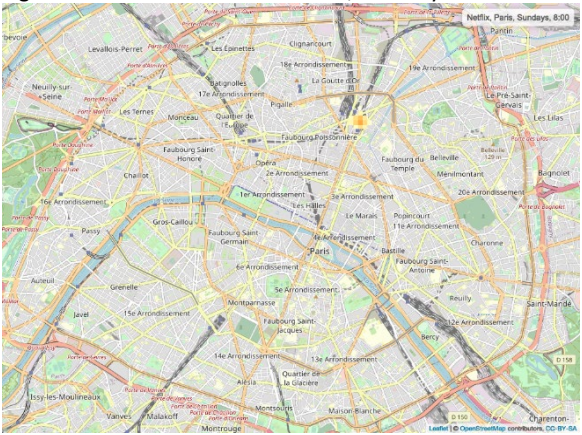

Fig.1f: Concentration of Netflix use, Mondays 19:00-20:00

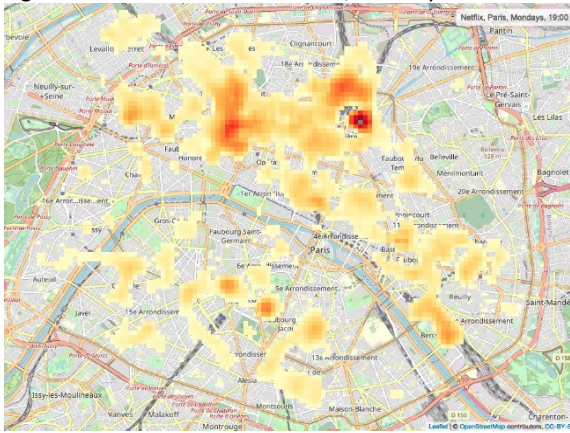

## 2.2 The ENACT population grid

ENACT (Batista e Silva et al., 2020) is a dataset of multitemporal population grids at a spatial resolution of 1 km x 1 km, capturing both intraday and seasonal population variations for the European Union (EU). The dataset is composed of 24 individual raster grid files covering the EU in a seamless fashion. The dataset was produced by combining two main data types: official statistics on population groups per subnational zoning systems and geospatial covariates of those population groups. The population grids of ENACT are based on monthly stocks of individual population groups at subnational level projected to grid-cell level using a population group-specific set of spatial covariates. The population groups included the number of residents, workers for different economic sectors, students, tourists, and non-working and non-studying population. The main stocks of population groups were obtained from official statistics. The monthly variations in population stocks were derived from school calendars as well as from monthly inbound and outbound tourists from official statistics. The population grids represent a typical working day of the month. However, the variation between workdays and weekend is not addressed. The nighttime frames represent an ideal situation assuming the whole population is at their place of residence or lodging to rest, whereas the daytime frames represent a situation whereby everybody is assumed to be at the location of their primary activity such as working or studying during core working hours.

For this application with the NetMob 2023 data, we used the population grid for France for April 2019. Since the night population is based on official statistics, so can be considered the ground truth. The day population corresponds to the distribution of the population across the grid and was estimated based on land use, land cover and Point-of-Interest data, and can thus be considered as a good approximation of the actual distribution of population during the day. Both population estimates were developed independently of the information included in the NetMob 2023 dataset and can therefore be considered as – in principle- suitable for modelling purposes.

The original ENACT grid uses a 1km x 1km resolution. In order to make it compatible with the NetMob 2023 100m x 100m grid, we applied a GIS procedure to downscale the ENACT grid and match it spatially with the NetMob 2023 grid. For the downscaling we applied the Akima algorithm (Akima, 1978) implementation from the R package (Akima et al., 2009). The downscaled population grid was then spatially joined to the grid of each of the 20 cities covered in NetMob 2023 using the sf package in R (Pebesma & Bivand, 2023).

Figures 2a and 2b visualize the distribution of the night and day population in the Greater Paris area that is covered in the NetMob 2023 dataset. While a certain mobility in the population is already visible, filtering for the areas that have a significant increase or decrease between night and day suggests that there is a strong trend for mobility between the outskirts of Paris to its centre and main activity areas (Figures 2c and 2d). In addition, it is worth mentioning that the total night population in the area covered is 7 million, which increases to 8 million during the day, an observation that indicates that there is a significant movement of population from/to zones outside the covered area.

The population grids are available for all 20 cities in NetMob 2023 and in practically all cases reveal a high level of mobility (Table 1). In most cases, the urban area population increases during the day, but in Marseille, Nice and Saint-Etienne it appears that the population movement is the reverse. The population figures also suggest that there is a high variance in terms of the population density among the 20 population grids. For example, the grid of Paris has a comparable area to that of Rennes, even though Paris has 17.5 times the population of Rennes.

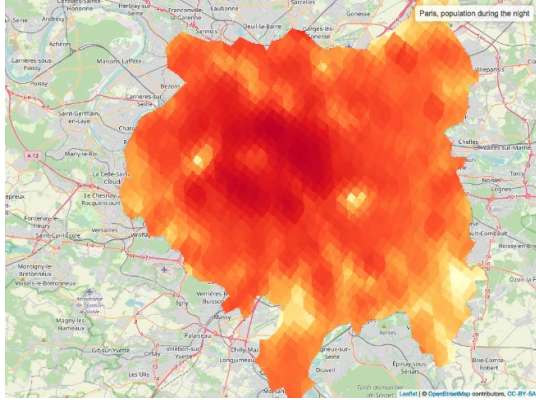
Fig. 2a: Distribution of population in Paris during the night

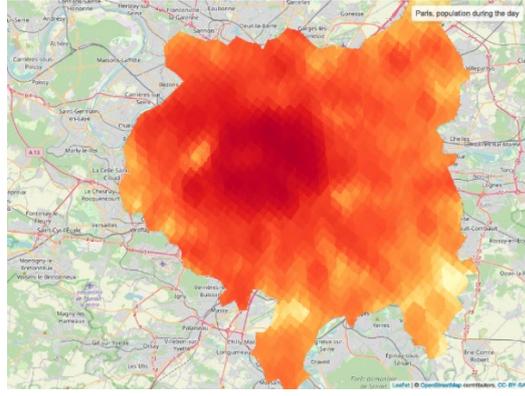
Fig. 2b: Distribution of population in Paris during the day

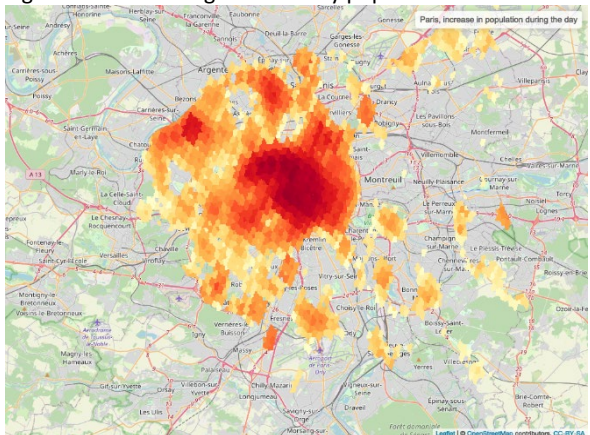
Fig. 2c: Areas with significant daily population increase

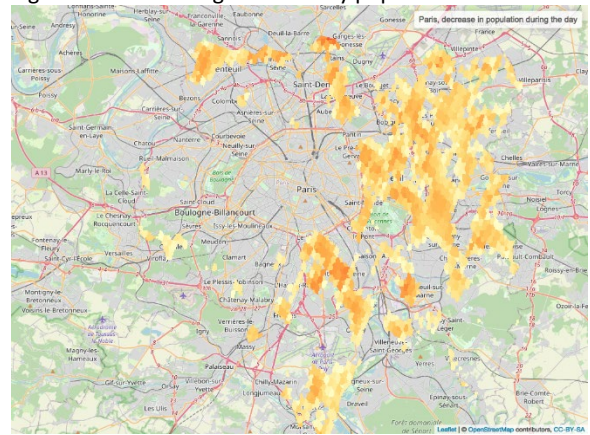
Fig. 2d: Areas with significant daily population decrease

source: ENACT dataset, converted by the authors to 100m x 100m grid

Table 1: Nighttime and daytime population in 20 urban areas covered in NetMob 2023 dataset

|  | night | day | % change day |
|---|---|---|---|
| Paris | 6974 | 8064 | 15.6% |
| Lyon | 1290 | 1445 | 12.0% |
| Lille | 1141 | 1213 | 6.3% |
| Marseille | 836 | 819 | -2.1% |
| Bordeaux | 716 | 821 | 14.6% |
| Toulouse | 705 | 786 | 11.5% |
| Nantes | 588 | 631 | 7.2% |
| Strasbourg | 471 | 488 | 3.6% |
| Nice | 455 | 436 | -4.2% |
| Grenoble | 431 | 442 | 2.5% |
| Montpellier | 419 | 491 | 17.1% |
| Rennes | 407 | 453 | 11.3% |
| Saint-Etienne | 397 | 387 | -2.6% |
| Tours | 283 | 315 | 11.3% |
| Clermont-Ferrand | 275 | 300 | 9.1% |
| Orleans | 268 | 289 | 7.9% |
| Nancy | 251 | 270 | 7.5% |
| Dijon | 244 | 271 | 11.1% |
| Metz | 223 | 242 | 8.7% |
| Mans | 199 | 236 | 18.4% |

## 3. Modelling

The core of our methodology consists of using the NetMob 2023 data as independent variables to predict the population distribution of each city during the night and during the day. The main hypothesis is that the use of each mobile service in space and time is proportional to the population present in the area when the temporal patterns of activity are also taken into account:

$$P_i^\pi = \sum_{a \in A} \sum_{d \in D} \sum_{t \in T} (w_{adt}^\pi M_{adt}^i) + \varepsilon_i^\pi$$

(Eq. 1)

where:

$P_i^\pi$ is the population in tile i during period $\pi$

$a$ is a specific service provider from a set of A service provider

d as specific day among a set of D days used in the model

t is a specific time slot within day d, among a set of T days used in the model

$M_{adt}^i$ is the data traffic for service provider $a$ in tile i during day d and time slot t

$w_{adt}^\pi$ is the weight given to the values of M for each a, d, t combination

$\varepsilon_i^\pi$ an error term

Our approach consisted of developing 3 sets of models that predict population per 100mx100m cell using the data for each of the 68 services in the dataset.

Night population model:

$$P_i^{night} = \sum_{a \in A} \sum_{d \in D} \sum_{t \in T} (w_{adt}^{night} M_{adt}^i) + \varepsilon_i^{night}$$

(Eq. 2)

Day population model:

$$P_i^{day} = \sum_{a \in A} \sum_{d \in D} \sum_{t \in T} (w_{adt}^{day} M_{adt}^i) + \varepsilon_i^{day}$$

(Eq. 3)

Day population model with known night population:

$$P_i^{day} = P_i^{night} + \sum_{a \in A} \sum_{d \in D} \sum_{t \in T}(w_{adt}^{daynight} M_{adt}^i) + \varepsilon_i^{daynight}$$

(Eq. 4)

The NetMob data we used is the average per weekday/ 2 hour time slot, a level of aggregation that leads to satisfactory precision levels, while still allowing a manageable dataset size. We generated the model features (independent variables) by converting the NetMob 2023 to a wide format. For each tile i in the dataset, we estimated the mean data traffic for each service provider $a$, for an average day of the week, traffic direction (upload or download) and time slot. This resulted in 68x7x2x12= 11424 features (columns) to be used as independent variables.

We explored the impact of various independent variables through a tree-based regression model. Such a model allows the identification of the factors with the highest impact on the dependent variable, allowing non-linear relationships and variable co-linearity to be taken into account. The specific algorithm used was Xgboost (Chen & Guestrin, 2016), a well-known gradient-boosting library that provides highly accurate results. We followed the standard machine learning approach of splitting the data in train, test and validation subsets (40%, 40% and 20% respectively).

The initial tests demonstrated a high correlation in traffic values from Monday to Thursday. In order to make the problem less computationally challenging, we replaced the values for those four days with their means and, as a result used only 4 options for d: (Friday, Saturday, Sunday, Weekday). This change led to the number of features decreasing to 68x4x2x12=6528.

Once the optimal model configuration was reached, we measured the importance of each model variable using SHAP values (SHapley Additive exPlanations), a tool that allows transparency and interpretability of machine learning models (Liu & Just, 2019). SHAP uses a game theoretic approach to explain the output of any machine learning model. It connects optimal credit allocation with local explanations using the classic Shapley values from game theory. The SHAP value analysis shows the contribution or the importance of each feature on the prediction of the model and allows a better understanding of how each underlying variable influences the predicted population in each tile.

## 4. Results

The 3 models were developed and extensively tested with the NetMob 2023 data available for the Paris urban area and were replicated for the other urban areas included.

The first model, predicting night population achieved quite high precision levels when comparing the modelled population at grid level with the ground truth (RMSLE= 0.243). Figure 3 shows the top-20 features that have an impact on the prediction. The one with the highest impact is the average volume of uploads for Spotify on Fridays during the 22:00-00:00 time slot, followed by the average volume of downloads for Google Meet on Sundays during the 20:00-22:00 time slot. The sum of downloads for all services on a typical weekday from 0:00 to 02:00 is also a rather important indicator of population. The top 20 features provide 77.3% of the total model gain, while the top 100 features provide 87.0% of gain. Most of the top features correspond to time slots during the night, which is quite intuitive. Nevertheless, the model still uses features that correspond to time slots during the day since they also contribute, even though with a lower weight.

Fig. 3: Model feature importance for night population model, Paris

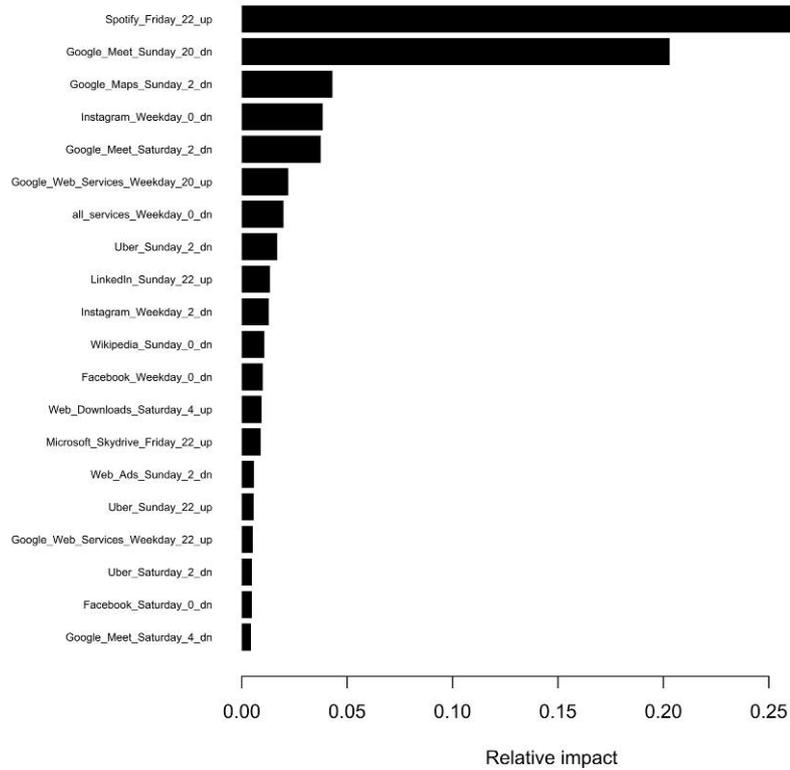

Fig. 4: SHAP dependence graphs, Paris night model:   Spotify uploads Fridays 22:00-00:00 (left)
                                                    Google Meet downloads Sundays 20:00-22:00 (right)

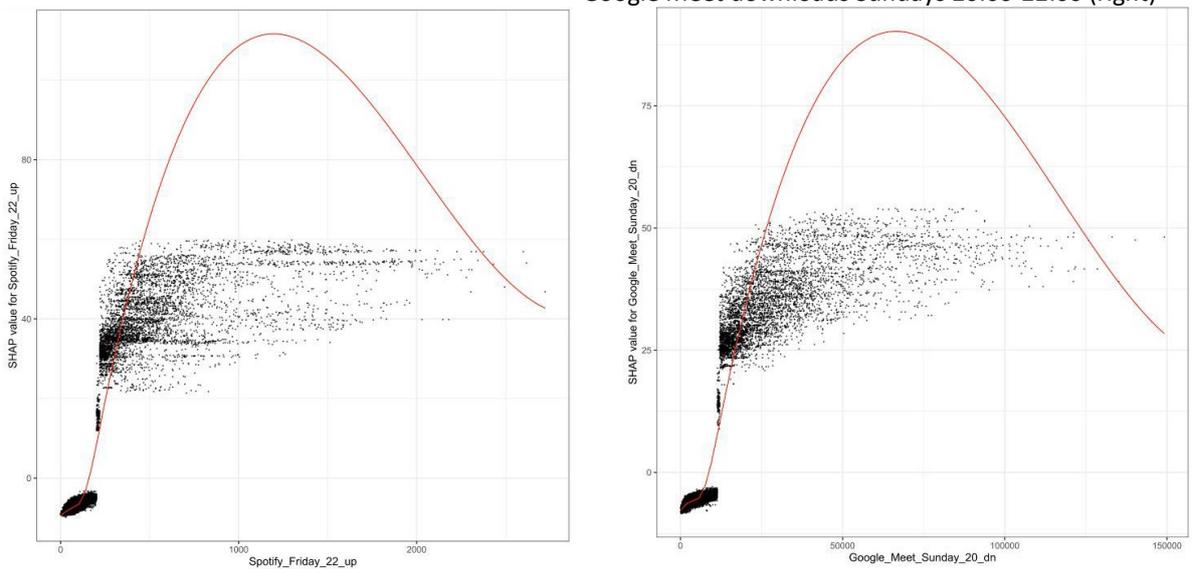

The SHAP dependence graph for the two most important variables (Fig. 4) shows that there is a strong correlation between their respective data traffic volumes and the level of a tiles population, as well as a clear cut-off level under which the model decreases the population predictions. The SHAP values also reveal the high variance in the use of each service among cells with similar population. Those two observations are useful for the understanding the structure of the –in essence tree-based- model and the implications on the value of the NetMob 2023. Each feature used provides additional value to the model by allowing the observed data to be partitioned. Nevertheless, each individual feature entails has a high variation and cannot be sufficient to explain the target variable on its own. Combining a large number of such features increases the overall precision of the model considerably.

The day population model gave higher accuracy levels (RMSLE= 0.196). This suggests that the NetMob 2023 data do an even better job in predicting day population than night population, even though the underlying data are not derived from census data and may entail higher uncertainty levels. This can be considered as a positive attribute for the NetMob 2023 data in terms of their suitability to capture daily mobility patterns. The top features in terms of importance show a few changes. Spotify uploads on Fridays 22:00-00:00 are still high in the list, but Google Maps downloads now occupy the top 3 positions. While the weight of Google Maps downloads now occupies the top 3 spots. The top-20 list includes more features that correspond to daytime time slots, but activity at night time still remains relevant. The concentration of the relative weight in the top features is higher in this case than in the night model. The top 20 features provide 84.0% of model gain, while the top 100 features provide 91.6% of model gain.

Fig. 5: Model feature importance for day population model, Paris

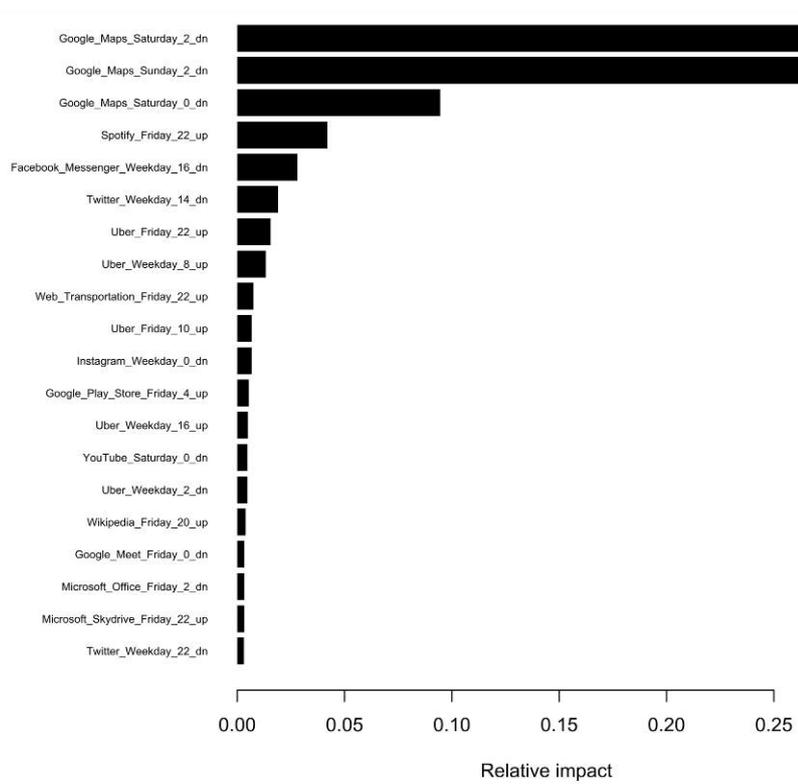

The SHAP dependence graphs for the Paris day population model follow, in principle, a similar pattern with those in the night model. But, for example, Google Maps downloads on Saturdays 02:00-04:00 (Fig. 6) has two different cut-off points with differing levels of variance.

Fig. 6: SHAP dependence graph, Paris day model: Google Maps downloads, Saturdays 02:00-04:00

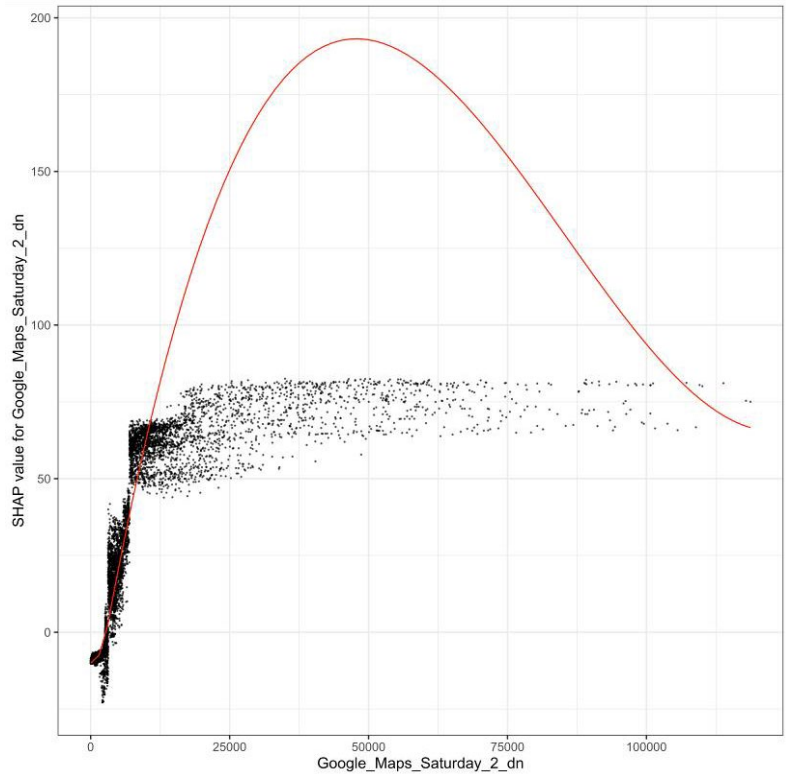

The third model for Paris estimates daytime population for each tile using the night population data as an additional input on top of the NetMob 2023 data. This addition leads to a significant further increase in the model precision, with RMSLE decreasing further to 0.110. This improvement is clearly the result of the inclusion of night population as input to the model as reflected in the list of the top-20 features (Fig.7). Night population is indisputably the feature that contributes to the model's impact the most, with more than 70%. Given this high value, it is not surprising that the top 20 features contribute to 95.0% of model gain and the top 100 to 97.8% of model gain.

The SHAP dependence graph for night population (Fig. 8) show a high correlation between day and night population at tile level, with the variance increasing for tiles with higher night population.

Fig. 7: Model feature importance for day population model using night population as input, Paris

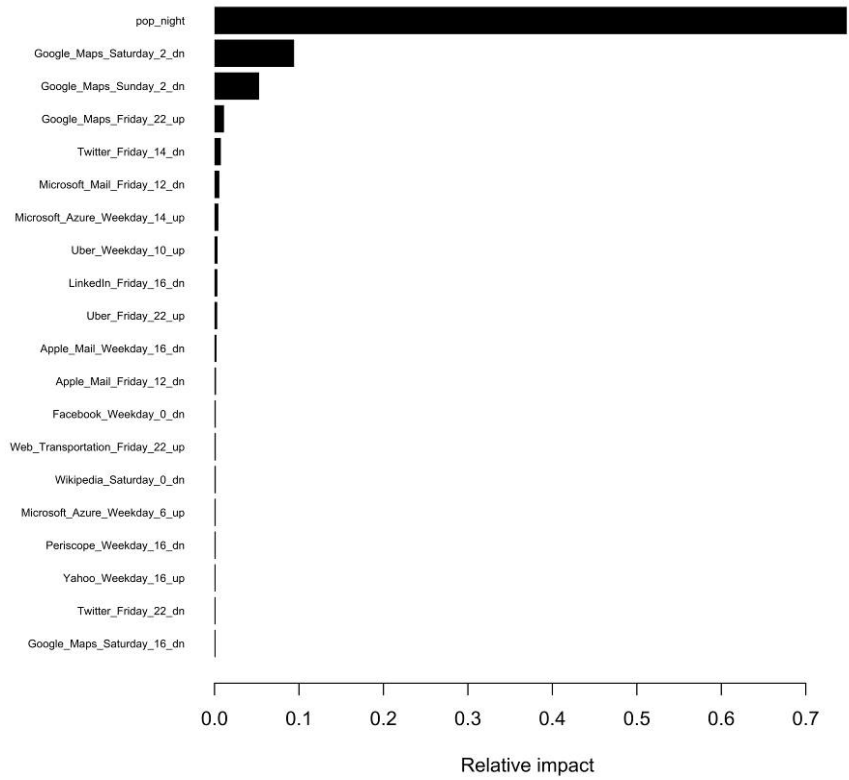

Fig. 8: SHAP dependence graph, Paris day model using night data as input: night population

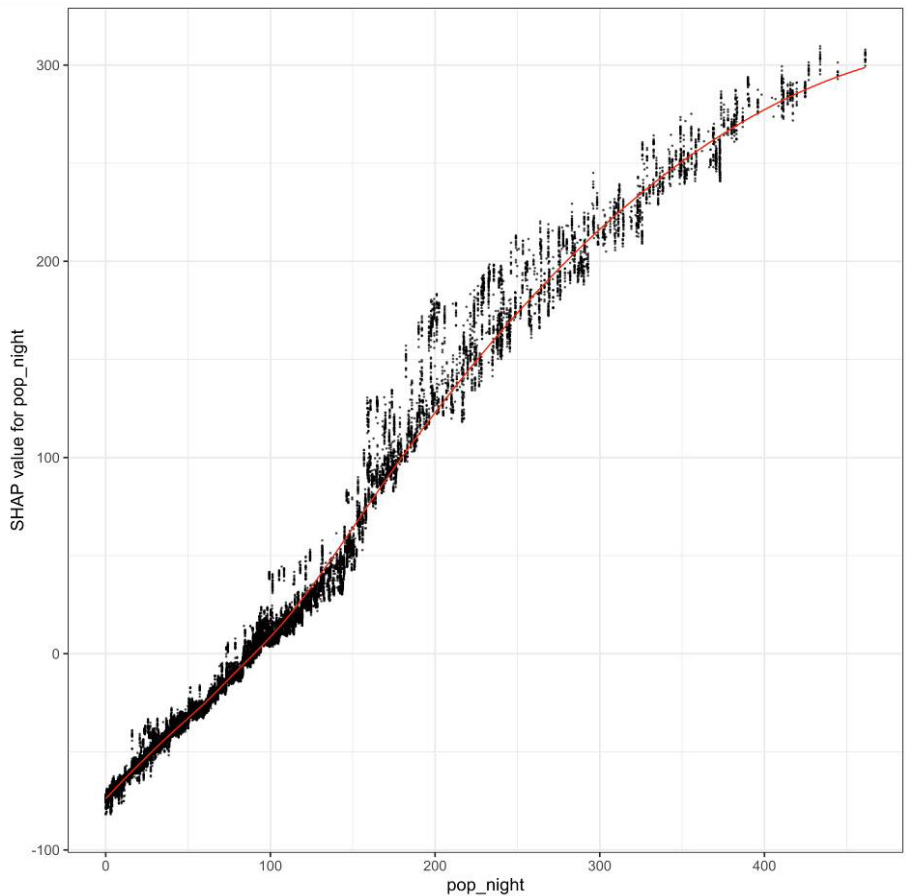

A visualization of the top features on a map can help in interpreting the importance of each variable and evaluating the usefulness of the data (Fig. 9-13). The tiles with higher concentration would be considered in the model as more likely to have a higher population during the day than the ones with lower concentration. The combined effect of using all 6528 features plus the night time population with their respective weights can provide a good approximation of the day time population in each tile.

Fig. 9: Distribution of data traffic for Google Maps downloads on Saturdays 02:00-04:00, Paris

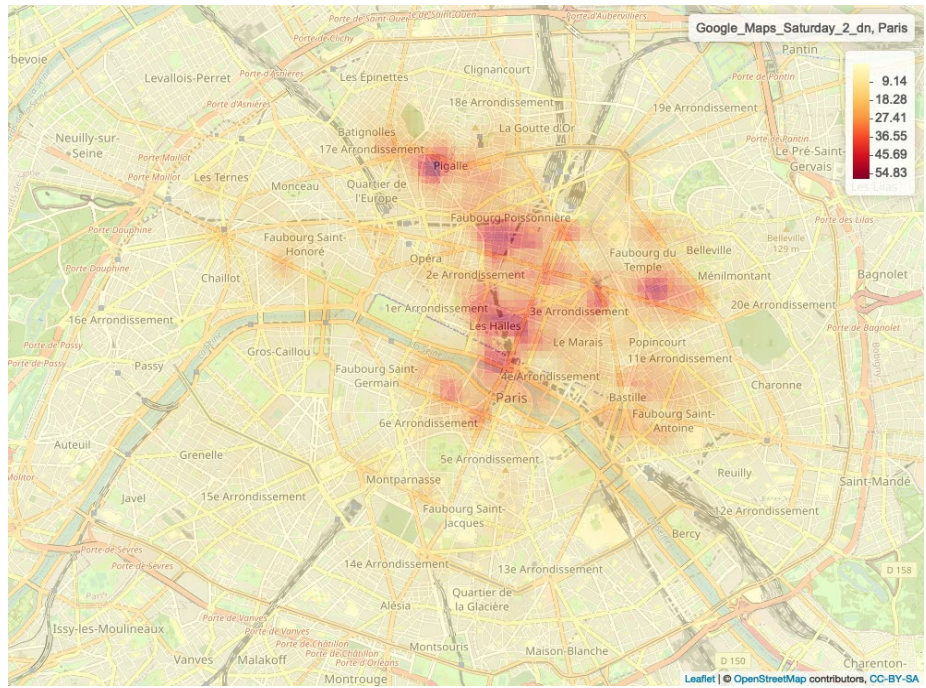

Fig. 10: Distribution of data traffic for Google Maps downloads on Sundays 02:00-04:00, Paris

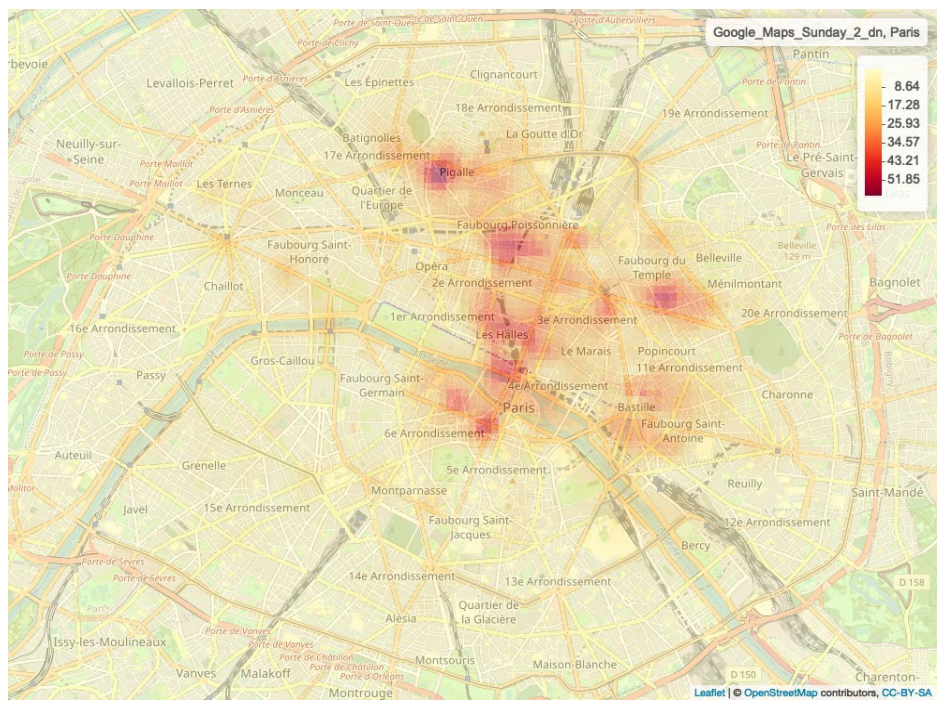

Fig. 11: Distribution of data traffic for Google Maps uploads on Fridays 22:00-00:00, Paris

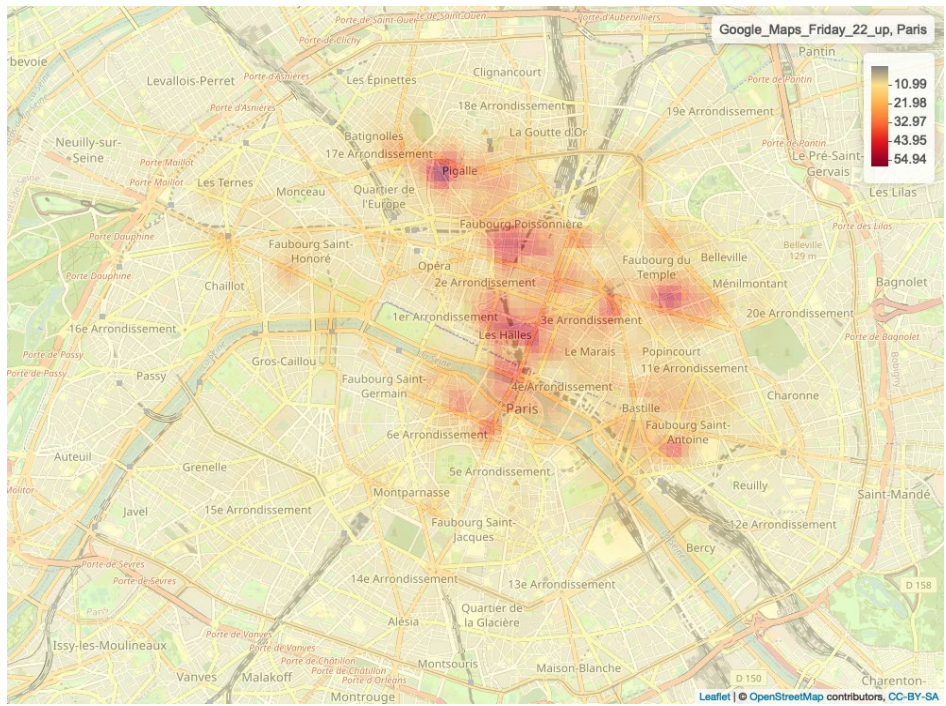

Fig. 12: Distribution of data traffic for Twitter downloads on Fridays 14:00-16:00, Paris

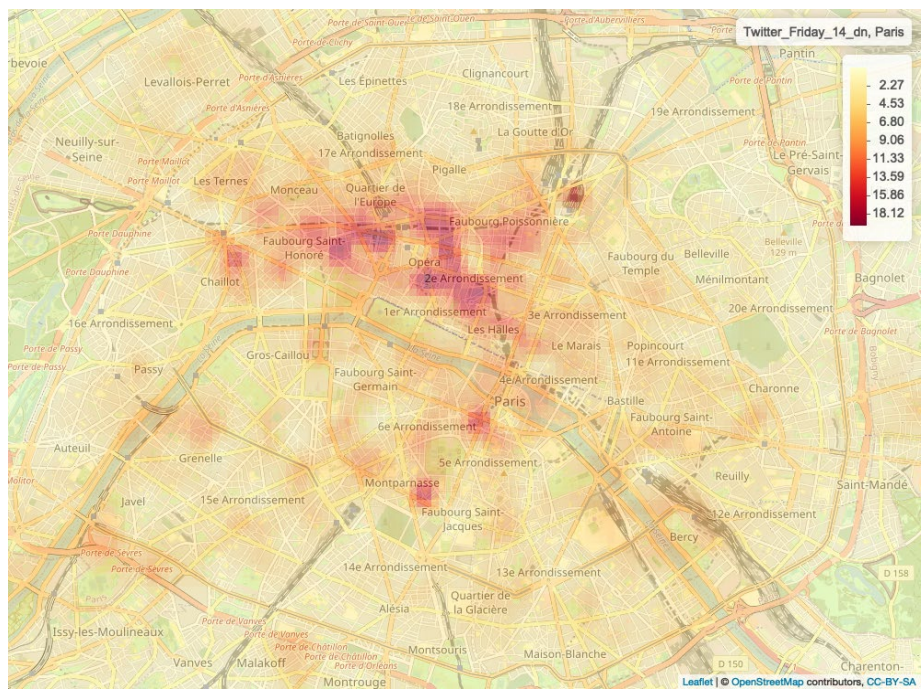

Fig. 13: Distribution of data traffic for Microsoft Mail downloads on Fridays 12:00-14:00, Paris

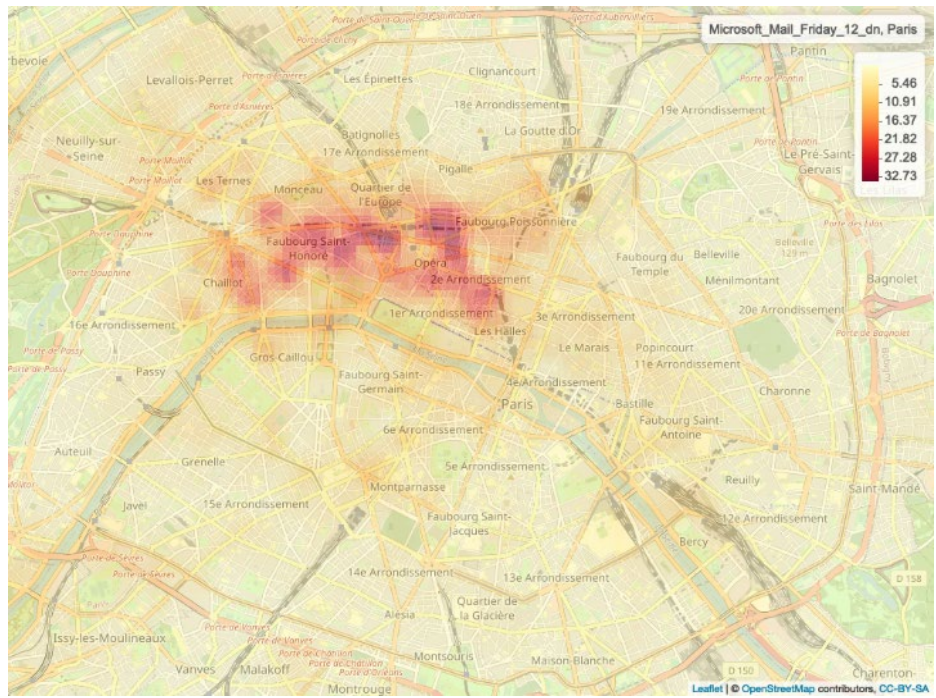

An other important aspect that we explored was the transferability of the results to other urban areas. Following a similar approach in Dijon and Marseille gave acceptable results, but did not achieve comparable levels of accuracy (RMSLE= 0.306 for Dijon and 0.305 for Marseille). The list of the top 20 in terms of impact for the two cities (Fig. 14 and 15) differs significantly between them, as well as with Paris. The contribution of night population is visibly lower in both cases than in Paris.

Fig. 14: Model feature importance for day population model using night population as input, Dijon

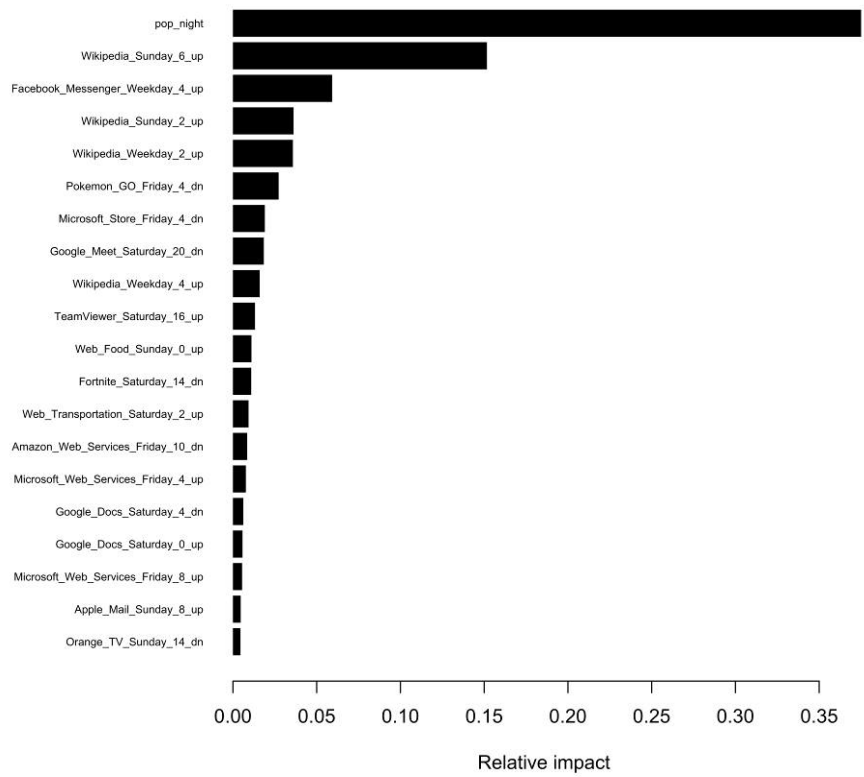

Fig. 15: Model feature importance for day population model using night population as input, Marseille

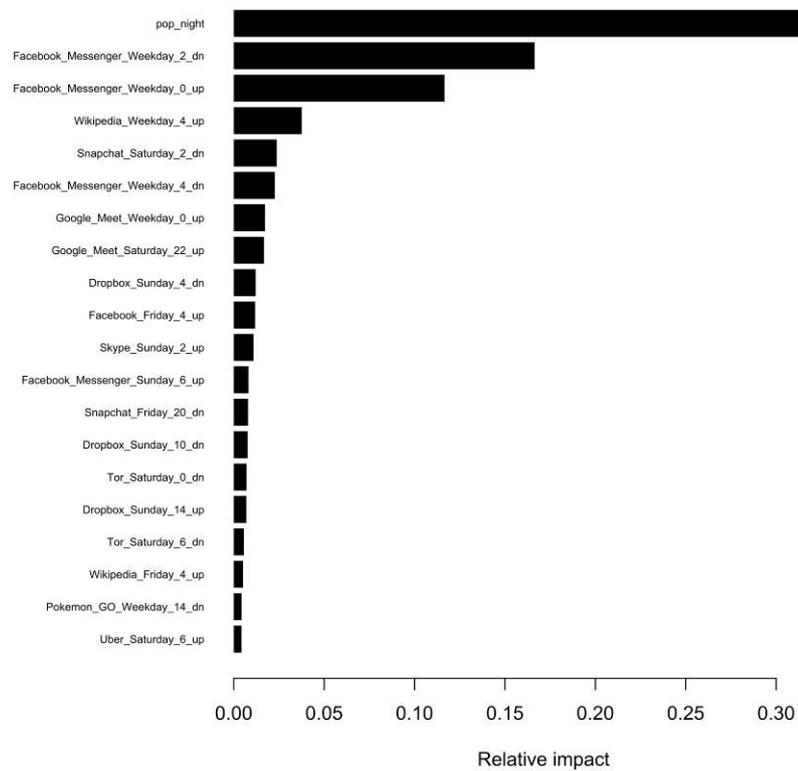

## 5. Discussion and conclusions

The results presented here summarize the progress made with the analysis of the NetMob 2023 data within the time constraints of the Data Challenge. While some preliminary conclusions can be drawn, they are in no case definitive, since further tests and analyses should be carried out.

The overall hypothesis that we tested, whether the mobile data traffic provided is useful for capturing human mobility in urban areas, seems to hold. We developed an XGBoost model that can predict a set of population indicators at 100m x 100m grid cell level using the NetMob 2023 data. The model accuracy is satisfactory for all three model versions (night, day, day using night population as input) and the weights of the main features are rather plausible. Additionally, the model performs better in the prediction of day population than for night population, suggesting that the data are suitable for the analysis of urban mobility.

There are, however differences in the implementation across cities. The relative weight of the features changes significantly between cities. This can be due to different population profiles, mobile phone and application use patterns, which could explain differences in the average data traffic generated by residents and visitors in each city. The city profiles in the sense of inflows and outflows also seems to play a role. The number and distribution of visitors to the city during the day varies among the 20 cities and can lead to different weights for each feature. Population density and personal daily mobility also differ depending on the city size and profile. The reliability of the underlying data, both from NetMob 2023 and from ENACT, may also be different and can affect the overall results.

We applied a specific XGBoost approach, using the mean values of the original data for 2-hour slots for specific days of the week. This allowed a manageable size for the model data and drastically limited the noise in the observations. This approach for feature engineering is suitable for the task that we were addressing, the estimation of the average population in each tile, but certainly sacrifices some quality in the data. Future work can address different model configurations and/ or algorithms, especially those that allow the exploration of finer temporal granularity. We did exploit the spatial granularity at its maximum (100m x 100m) and the results are satisfactory. An alternative to explore the granularity/applicability trade-off could be to use a coarser spatial scale (e.g. 500m x 500m) in combination with a finer temporal scale.

While we followed the standard machine learning techniques to ensure the robustness of the approach (data splitting, cross-validation, model testing and fine-tuning), there is always room for further improvements. In future work, we will further explore the impacts of collinearity among the features used and will test additional model hyper-parameters.

To summarize, the approach presented here is a first, promising step for the use of the NetMob 2023 data in analyzing urban dynamics and human mobility. Our results demonstrate that the data can provide added value and can be useful for several applications.

Acknowledgement: The authors would like to thank Orange and the NetMob 2023 Data Challenge organisers for the provision of the dataset